\documentclass[sigconf]{acmart}
\AtBeginDocument{%
  }

\acmConference[K-CAP 2025]{International Conference on Knowledge Capture}{December 10--12,
  2025}{Ohio, USA}

\usepackage{graphicx}
\usepackage{subcaption}
\usepackage{caption}
\usepackage{multirow}
\usepackage{graphicx}
\usepackage[table,xcdraw]{xcolor}
\usepackage{algorithm} 
\usepackage{algorithmicx} 
\usepackage[noend]{algpseudocode}
\begin{document}

\title{Improving Knowledge Graph Embeddings through \\ Contrastive Learning with Negative Statements}

\author{Rita T. Sousa}
\email{rita.sousa@uni-mannheim.de}
\orcid{0000-0002-7241-8970}
\affiliation{%
  \institution{Data and Web Science Group, University of Mannheim}
  \city{Mannheim}
  \country{Germany}
}

\author{Heiko Paulheim}
\email{heiko.paulheim@uni-mannheim.de}
\orcid{0000-0003-4386-8195}
\affiliation{%
  \institution{Data and Web Science Group, University of Mannheim}
  \city{Mannheim}
  \country{Germany}
}

\renewcommand{\shortauthors}{Sousa et al.}

\begin{abstract}
Knowledge graphs represent information as structured triples and serve as the backbone for a wide range of applications, including question answering, link prediction, and recommendation systems. 
A prominent line of research for exploring knowledge graphs involves graph embedding methods, where entities and relations are represented in low-dimensional vector spaces that capture underlying semantics and structure. 
However, most existing methods rely on assumptions such as the Closed World Assumption or Local Closed World Assumption, treating missing triples as false. This contrasts with the Open World Assumption underlying many real-world knowledge graphs. Furthermore, while explicitly stated negative statements can help distinguish between false and unknown triples, they are rarely included in knowledge graphs and are often overlooked during embedding training.

In this work, we introduce a novel approach that integrates explicitly declared negative statements into the knowledge embedding learning process. Our approach employs a dual-model architecture, where two embedding models are trained in parallel, one on positive statements and the other on negative statements. During training, each model generates negative samples by corrupting positive samples and selecting the most likely candidates as scored by the other model. The proposed approach is evaluated on both general-purpose and domain-specific knowledge graphs, with a focus on link prediction and triple classification tasks. The extensive experiments demonstrate that our approach improves predictive performance over state-of-the-art embedding models, demonstrating the value of integrating meaningful negative knowledge into embedding learning.  

\end{abstract}



\maketitle

\section{Introduction}

A knowledge graph (KG)~\cite{hogan2021knowledge} is a structured representation of knowledge that encodes information as a set of triples $(h, r, t)$. For each triple, the head $h$ is an entity, the tail $t$ is either an entity or a literal (represented as nodes), and the relation $r$ (represented as an edge) describes the relationship or property connecting them.
Over the years, public and commercial KGs have been developed~\cite{vrandevcic2014wikidata}, reflecting the increasing importance of structured knowledge in diverse domains. 
One of the key strengths of KGs is their ability to be understood by both humans and machines, allowing them to support a wide range of applications, including question answering~\cite{huang2019knowledge}, link prediction~\cite{rossi2021knowledge}, and recommendation systems~\cite{guo2020survey}. 

A variety of approaches have been developed for KG-based tasks. Among these, KG embedding (KGE) methods have gained significant attention, with a wide range of methods proposed in the literature. KGE methods encode entities and relationships as low-dimensional vectors within a latent embedding space, aiming to capture and preserve the underlying structural information and graph properties~\cite{wang2017knowledge}. 
The majority of these methods learn embeddings by defining a scoring function that measures the likelihood of a given triple being true. During training, the model adjusts the embeddings to assign higher scores to triples that exist in the KG and lower scores to negative triples. The process of generating negative triples, known as negative sampling, typically involves corrupting true triples by randomly replacing their head or tail entity with another entity from the KG~\cite{hubert2024treat}. 

Although most KGs are built and interpreted under the Open World Assumption (OWA), which means that the non-stated facts may represent unknown facts or true negative statements, many KGE models adopt the Closed World Assumption (CWA), or more commonly, the Local Closed World assumption (LCWA)~\cite{doherty2000efficient}, especially during the generation of negative triples. Under LCWA, if a triple $(h, r, t)$ exists, then the KG is considered to include all possible tails for any statements of the form $(h, r, ?)$. This works reasonably well for relations expected to have a single tail, such as $hasCapital$. However, it breaks down for relations that can link a head to multiple tails, like $worksAt$, which is typical for most real-world relations~\cite{galarraga2013amie}.
Recognizing these assumptions is critical for many KG-related tasks, namely link prediction. To predict new links accurately, it is first necessary to distinguish between links that are false and those that are merely unknown. This highlights the relevance of explicitly defining negative statements in the KG.

Negative statements have long been underrepresented in KGs, despite the growing recognition of their importance. They are either filtered out during construction or treated no differently than positive relations, merely encoded as another edge type. As a result, only some studies have explored how to incorporate negative knowledge into representation learning~\cite{sousa2023biomedical}. More recently, however, some research initiatives focus on enriching existing KGs with interesting and meaningful negative statements. Given the infinite pool of possible negative statements, the most valuable are typically those that people might reasonably expect or believe to be true, but which are ultimately shown to be false. Notable examples include efforts to enrich Wikidata with negative statements~\cite{arnaout2021wikinegata}, generate informative negative commonsense statements~\cite{arnaout2023uncommonsense}, or extend a widely used biomedical ontology with statements indicating that a protein does not perform a given function~\cite{vesztrocy2020benchmarking}.  

In this work, we address the gap left by most existing KGE methods by proposing an approach that incorporates explicitly declared negative statements and adapts the negative sampling process using a contrastive learning strategy. 
Our approach trains separate KGE models on positive and negative statements, and generates negative samples by corrupting triples and selecting the most plausible corrupted triples based on the scores from the opposing model.
We assess the effectiveness of our approach on different KGs, including a general-purpose KG like Wikidata and a domain-specific biomedical KG such as the Gene Ontology (GO). Our evaluation covers two tasks: link prediction and triple classification for protein-protein interaction (PPI) prediction. Our contributions are as follows:
\begin{enumerate}
\item The design of a dual-model training framework that learns separate embeddings for positive and negative statements.
\item A novel negative sampling strategy based on contrastive learning that iteratively refines negative samples, leading to more challenging negatives.
\item Extensive experiments to demonstrate the effectiveness, robustness, and broad applicability of the proposed approach on diverse KGs and tasks. 
\end{enumerate}

\section{Related Work}

There is a substantial body of work on KGEs, as highlighted by several surveys~\cite{wang2014knowledge}. 
Broadly, research in this area has evolved along two main directions~\cite{portisch2022knowledge}. The first focuses on creating embeddings of KG entities for downstream tasks, with an emphasis on capturing semantic similarity. Methods in this category, such as RDF2Vec~\cite{ristoski2016rdf2vec}, are typically based on natural language processing and random walks, and are evaluated on external ground truth data outside the KG. The second direction centers on link prediction, where the objective is to distinguish between correct and incorrect triples using a scoring function. This line of work includes translational distance models~\cite{bordes2013translating} or semantic matching models~\cite{yang2014embedding}. 

To the best of our knowledge, TrueWalks~\cite{sousa2023biomedical} is the only KGE-based approach that incorporates explicitly defined negative statements during embedding learning for ontology-rich KGs.
TrueWalks generates two distinct embeddings for each entity: one capturing the positive semantics and another capturing the negative semantics. For the positive embedding, it generates walks based on positive and $subClassOf$ relationships, then inputs these walks into a skip-gram model to learn the embedding. Similarly, for the negative embedding, it generates walks using negative and $superClassOf$ relationships, feeding them into a skip-gram model to obtain the negative embedding. This dual embedding strategy allows TrueWalks to account for the semantic implications of negation in ontology-rich KGs. Despite its effectiveness, TrueWalks has some limitations. First, its applicability is restricted to ontology-rich KGs, which are prevalent in the biomedical domain but less common in other areas. Additionally, by learning positive and negative embeddings independently, TrueWalks does not explicitly model how contradictory statements affect entity dissimilarity.

Beyond the KGE literature, some works have explored explicit negative information using graph neural networks (GNNs), particularly to model whether a user is or is not interested in an item in recommendation systems.
SiGAT~\cite{huang2019signed} uses social theories (balance and status theory) to categorize neighbor nodes into motifs. Then, it applies a GAT-based aggregation to combine information from those categorized neighbors.
SiGRec~\cite{huang2023negative} creates two separate embeddings per node (positive and negative) and combines them by concatenation. It also introduces the sign cosine loss, a loss function designed to handle various types of negative feedback. 
Lastly, PANE-GNN~\cite{liu2023pane} partitions the graph into two distinct bipartite graphs based on positive and negative feedback and then generates an interest embedding and a disinterest embedding with positive and negative edges. For the negative graph, a distortion is introduced to denoise the negative feedback. 

In summary, although prior work has explored negative information in KG representation, several gaps remain. 
GNN-based approaches are designed for homogeneous graphs and struggle with the heterogeneous, richly typed relations in KGs. As a result, directly applying such architectures to KGs often overlooks complex semantics.
KGE models, on the other hand, are capable of capturing such semantic information, but the only existing approach that explicitly handles negative statements is path-based and cannot be extended to link prediction approaches.
The proposed approach overcomes these limitations by training separate link prediction KGE models on positive and explicitly declared negative statements, using each to guide the other’s negative sampling.

\section{Methodology}

\begin{figure}[t!]
    \centering
    \includegraphics[width=0.6\linewidth]{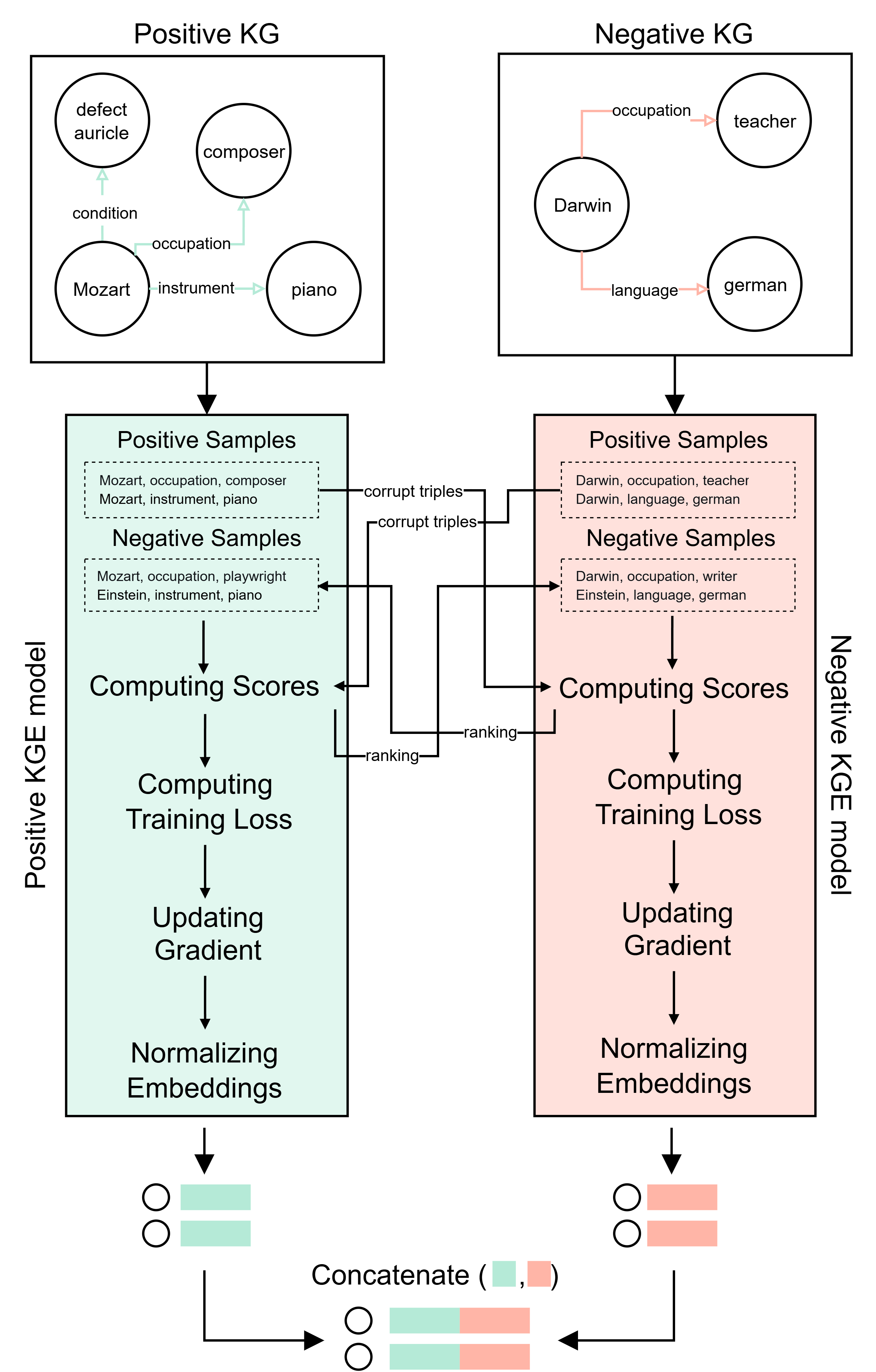}
    \caption{Overview of the proposed approach that trains two models, one on the positive KG and another on the negative KG, using each to guide the other’s negative sampling.}
    \label{fig:methodology}
    \vspace{-0.3cm}
\end{figure}

An overview of the proposed approach for generating contrastive negative samples is illustrated in Figure~\ref{fig:methodology}.
First, two separate KGs are built: one from positive statements and another from explicitly defined negative statements. For each KG, a KGE model is then initialized and trained. In standard KGE training, triples from the KG are used directly as positive samples while negative samples are generated through random corruption (e.g., replacing the head or tail entity of a triple with a random entity). In contrast, our approach introduces a contrastive learning strategy: for each triple in one KG, we replace either the head or the tail entity with all possible entities from the KG, then compute the plausibility scores using the model trained on the other KG. The corrupted triple with the highest score, which represents the most plausible according to the other model, is selected as the negative example. After training, each node in the KG is represented by the concatenation of its embeddings from both positive and negative KGEs, yielding a dual perspective that captures both positive and negative aspects. 

The proposed approach is KGE-agnostic and can be integrated into any KGE model that defines a scoring function and employs negative sampling during training. The implementation is available on GitHub\footnote{\url{https://github.com/ritatsousa/KGE_with_CL}}.

\subsection{Building the Positive and Negative Knowledge Graphs}

The positive and negative KGs are constructed as RDF graphs.
An RDF graph is a directed, labeled graph in which edges represent the named relations between two nodes, and are represented as triples in the form $(h, r, t)$. The nodes may correspond to individuals, classes, literals, or other RDF resources.
When the KG is backed by an ontology, the transformation follows the OWL to RDF Graph Mapping guidelines defined by W3C\footnote{https://www.w3.org/TR/owl2-mapping-to-rdf/}. Simple OWL axioms are directly translated into RDF triples. More complex axioms, especially those involving nested class expressions, are decomposed into multiple RDF triples and may include blank nodes to preserve structure and semantics.

\subsection{Initializing the Embedding Models}

Once the KGs are built, the KGEs models can be initialized to learn low-dimensional vector representations of entities and relations. 
Numerous KGE approaches have been proposed in recent years~\cite{wang2017knowledge}, particularly for link prediction tasks, to which our approach can be integrated. Despite their differences, these KGE models typically share common components, including the definition of a scoring function, embedding initialization, optimization strategy, and loss function.
The scoring function quantifies the plausibility of a given triple $(h, r, t)$. 
Training involves learning entity and relation embeddings by minimizing a loss function over several epochs.
The goal is to assign higher scores to positive samples (triples that exist in the KG) and lower scores to negative samples (triples that are absent from the KG and are presumed false). 
Standard loss functions include margin-based ranking loss or binary cross-entropy, and optimization is usually carried out using stochastic gradient descent or its variants. 

In this work, we focus on integrating the proposed approach into three widely adopted KGE models. 
TransE~\cite{bordes2013translating} is a translational model that represents relations as translation vectors in the embedding space. It assumes that for a triple $(h, r, t)$, the vector of the tail $t$ should be close to the vector of the head $h$ plus the vector of the relation $r$. DistMult~\cite{yang2014embedding} models the score of a triple using a bilinear dot product of the entity and relation embeddings. 
ComplEx~\cite{trouillon2016complex} extends DistMult into the complex vector space by incorporating both real and imaginary components in the embeddings. Table~\ref{tab:scoring-kge} summarizes their scoring functions.

\begin{table}[t!]
\scriptsize
\caption{Scoring functions for the employed KGE models.}
    \begin{tabular}{ll}
         \toprule
         \textbf{KGE} &  \textbf{Scoring Function}  \\ \midrule
             TransE & $f(h, r, t) = -\left\lVert \mathbf{h} + \mathbf{r} - \mathbf{t} \right\rVert_p$\\
             DistMult & $f(h, r, t) = \sum_{i} h_i \cdot r_i \cdot t_i$\\
             ComplEx & $f(h, r, t) = \operatorname{Re} \left( \sum_{i} h_i \cdot r_i \cdot t_i \right)$\\
    \bottomrule
    \end{tabular}
    \vspace{-0.3cm}
    \label{tab:scoring-kge}
\end{table}

\subsection{Contrastive Learning for Generating Negative Samples}

Generating negative samples is critical for KGE training~\cite{hubert2024treat}. 
The most widely adopted strategy is random negative sampling, where negative samples are created by corrupting positive triples, typically by replacing either the head or the tail entity with a randomly chosen entity from the KG. However, since the majority of the KGs reside under the OWA, such random corruption can often result in false negatives, as many triples that are not present in the KG may still represent valid facts. Additionally, replacing entities indiscriminately can yield trivial negatives that are semantically invalid.
More advanced sampling strategies have since been proposed, such as (i) replacing heads or tails of positive triples with entities seen in the same relation~\cite{socher2013reasoning}, (ii) selecting replacement entities of the same type as the original~\cite{krompass2015type}, (iii) assuming one relation per entity pair and picking entities connected through different relations than the current one~\cite{kotnis2017analysis}, (iv) using a pre-trained embedding model to select nearest neighbors of entities in vector space as negatives~\cite{kotnis2017analysis}, or (v) using the highest scoring triples of a model's previous iteration which are not contained in the KG as negatives~\cite{shaban2024snape}. 
These advanced sampling strategies incorporate entity type constraints to generate harder negatives, yet they still fall short in addressing the challenges posed by the OWA.
Beyond the KGE domain, negative sample generation has also been explored in active learning, where various strategies, such as contrastive learning–based methods~\cite{margatina2021active}, are employed to selectively acquire informative examples that enhance model training efficiency.

In the proposed approach, we employ a two-stage approach to negative sampling. In the initial phase of training, negative samples are generated using the standard random corruption strategy. However, after a predefined number of epochs, denoted as $cl\_phase$, we switch to a contrastive learning-based strategy (see Algorithm~\ref{alg:CL}).
In the contrastive phase, for each positive triple from one KGE model, we generate candidate negative triples by replacing either the head or the tail entity with every other entity in the KG. Each candidate triple is then scored by the KGE model trained on the other KG. For instance, given the triple $(Mozart, occupation, composer)$ in the positive KG, we replace the tail entity with all other entities to produce candidates such as $(Mozart, occupation, playwright)$ or $(Mozart, occupation, scientist)$. The negative model scores all candidate triples, and the one with the highest score (i.e., the most plausible according to the model trained on the explicit negative statements) is selected as the negative sample for the positive model.

The proposed contrastive learning strategy enables the dynamic generation of negative samples that are more meaningful and challenging as training progresses, thus improving the KGE model's ability to learn robust representations under the OWA. 

\begin{algorithm}[t!]
\scriptsize
\algrenewcommand\algorithmicindent{1.5em}%
\caption{Training strategy for the positive KG ($kg\_pos$) and the negative KG ($kg\_neg$), incorporating a contrastive learning phase. Initially, both models (one trained on the $kg\_pos$, the other on the $kg\_neg$) generate random negative samples. After a predefined epoch ($cl\_phase$), each model generates contrastive negative samples by scoring entity perturbations using the other model.}

\begin{algorithmic}[1]

\vspace{0.2em}

\Function{train}{$kg\_pos, kg\_neg$}  

    \State $epochs \gets$ number of training epochs\;
    \State $cl\_phase \gets$ epoch to start contrastive learning\;
    \State $pos\_model.\Call{init}{kg\_pos}$\;
    \State $neg\_model.\Call{init}{kg\_neg}$\;

    \For{$epoch$ in $epochs$}
        \If{$epoch > cl\_phase$}
            \State $pos\_model.\Call{get contrastive samples}{kg\_pos,neg\_model}$
            \State $neg\_model.\Call{get contrastive samples}{kg\_neg,pos\_model}$
        \Else
            \State $pos\_model.\Call{get random negative samples}$
            \State $neg\_model.\Call{get random negative samples}$
        \EndIf
        \State $loss\_pos \gets pos\_model.\Call{train step}$
        \State $loss\_neg \gets neg\_model.\Call{train step}$
    \EndFor
\EndFunction

\vspace{0.2em}

\Function{get\_contrastive\_samples}{$self,kg,contr\_model$}

    \State $ents \gets$ \Call{gets unique entities}{kg}\;
    \State $contr\_negative\_samples \gets []$\;
    \ForAll{$(h, r, t)$ in $kg$}
        \State $perturb\_heads\_flag \gets$ random flags for head/tail perturbation
        \If{$perturb\_heads\_flag$ is True}
            \ForAll{$candidate\_h$ in $unique\_ents$}
                \State $score \gets \Call{compute score}{candidate\_h, r, t, contr\_model}$
            \EndFor
            \State $best\_h \gets$ candidate with highest score
            \State $negative\_samples$.append($best\_h, r, t$) 
        \Else
            \ForAll{$candidate\_t$ in $unique\_ents$}
                \State $score \gets \Call{compute score}{h, r, candidate\_t, contr\_model}$
            \EndFor
            \State $best\_t \gets$ candidate with highest score
            \State $contr\_negative\_samples$.append($h, r, best\_t$) 
        \EndIf
    \EndFor
    \State $self.negative\_samples \gets contr\_negative\_samples$
\EndFunction

\vspace{0.2em}

\end{algorithmic}
\label{alg:CL}
\end{algorithm}

\subsection{Generating the Final Representation}

After training the positive and negative KGE models simultaneously, each node and relation in the KG is associated with two distinct representations: one learned from positive statements and another learned from negative statements. As already shown in~\cite{sousa2023biomedical} and~\cite{huang2023negative}, these dual perspectives capture complementary semantic information. To construct the final entity representation, we combine these two embeddings using vector concatenation.

\section{Evaluation}

We evaluate the proposed approach on two KGs: Wikidata KG, a general-purpose KG covering a broad range of domains, and the GO KG, a biomedical domain-specific KG focused on representing protein functions. Both KGs are enriched with negative statements and are described in the following sections. 

The evaluation tasks differ for each KG. For Wikidata, we perform link prediction, aiming to predict missing links. In contrast, for the GO KG, we perform triple classification, specifically framed as predicting whether a protein interacts with another protein. In both tasks, our approach generates lower-dimensional representations for KG entities. For link prediction on Wikidata, these embeddings are used to compute scores for candidate triples, which are then ranked to identify new links. For the PPI prediction task on the GO KG, it is cast as a binary classification problem, where we combine the embeddings of the entities and relation to create a composite representation of the triple, which is then fed into a supervised classifier to determine whether the interaction is likely to exist. 

As a baseline, we compare our approach against KGE models trained separately on the positive and negative statements. To ensure a fair comparison, the baseline models are trained using embeddings with twice the dimensionality of those in the proposed approach. This setup guarantees that the final embedding sizes of both methods are equivalent.

\subsection{Wikidata Knowledge Graph}

Wikidata~\cite{vrandevcic2014wikidata} is a free collaborative knowledge base supported by a large community of active users as well as automated bots. It contains a vast collection of statements describing millions of entities, with each entity typically associated with multiple statements.
The majority of these statements are positive assertions under the OWA. Some exceptions, such as deleted or deprecated statements, may imply negative information. Examples of positive statements include assertions like $(Barack\_Obama, occupation, Politician)$. Regarding its taxonomy, although it is built through collective efforts, the community faces challenges in deciding between $instanceOf$ or $subclassOf$ to model inheritance in Wikidata~\cite{shenoy2022study}.

Enriching Wikidata with negative statements has been addressed through a statistical inference method called peer-based inference~\cite{arnaout_negative_2021}.
The core idea of this method is to identify, for an entity $e$, a group of closely related entities referred to as peers. These peers provide a context from which negative statements about the entity $e$ can be formulated. The approach assumes that if a certain attribute is consistently present for most peers but missing for the entity $e$, the absence is interpreted as a negative fact rather than incomplete data. For example, if several U.S. presidents in the knowledge base are recorded as having a legal background or having worked as lawyers, but Barack Obama is not listed with this occupation, it is assumed that it is false for Obama and not a missing statement. Thus, the method infers the candidate negative statement $(Barack\_Obama, occupation, Laywer)$. 
A learning-to-rank model then scores candidate negative statements using features like predicate and object prominence, frequency, and textual context.
These negative statements can be found on the Wikinegata platform\footnote{https://d5demos.mpi-inf.mpg.de/negation/index.html}.

To construct the link prediction dataset, we filter the positive statements to retain only those involving entities that are also present in the set of negative statements. Formally, we include a positive triple $(h, r, t)$ in our dataset only if entity $h$ participates in relation $r$ in at least one negative statement, thereby ensuring that each entity is represented with both positive and negative statements. Subsequently, we split the positive statements into training and test sets, so that no entity or relation type is exclusive to the test set.
The resulting dataset comprises 743 798 positive statements and 453 325 negative statements in the training set, as well as a test set with 1 251 positive statements used to evaluate link prediction performance through ranking metrics.

\subsection{Gene Ontology Knowledge Graph}

GO KG integrates both the GO itself and the GO annotation data~\cite{gene2021gene}.
GO is structured as a directed acyclic graph, where nodes represent GO classes that describe gene product functions, and edges capture semantic relationships such as $isA$ or $partOf$. For example, the triple $(GO\_0055089, isA, GO\_0055088)$ expresses that fatty acid homeostasis (GO\_0055089) is a subclass of lipid homeostasis (GO\_0055088).
In addition to the ontology, GO annotations provide associations between gene products and GO classes, linking biological entities to their functional descriptions. In triple form, this can be represented as $(Insulin, hasFunction, GO\_0055089)$, indicating that insulin is involved in fatty acid homeostasis (GO\_0055089).

Recent efforts have enriched the GO KG by incorporating biologically meaningful negative GO annotations, which explicitly indicate that a protein does not perform a particular function. 
In this work, we employ the negative GO associations introduced by~\cite{vesztrocy2020benchmarking} derived from phylogenetic analysis of protein families. By tracking the evolutionary history of proteins across species, the phylogenetic trees reveal where functions are gained or lost over time. This enables the identification of functions that are absent in specific lineages and thus suitable to be recorded as negative annotations. For instance, if proteins from the insulin family once exhibited D-glucose transmembrane transporter activity but lost this function, the corresponding negative annotation can be represented as the triple $(Insulin, hasFunction, GO\_0055056)$, where GO\_0055056 denotes D-glucose transmembrane transporter activity. As a result, the GO KG comprises 91 701 positive statements derived from GO and GO annotations and 32 959 negative statements reflecting the negative GO annotations. 

Predicting PPIs is essential for understanding cellular processes and disease mechanisms, yet experimentally determining these interactions is expensive and time-consuming. Consequently, computational prediction methods aim to rank protein pairs based on their likelihood of interaction, thereby guiding experimental efforts effectively. Namely, functional information represented in the GO KG has been widely explored to predict PPIs in recent studies~\cite{sousa2020evolving}. 
We evaluate our approach using a PPI dataset introduced by~\cite{sousa2023benchmark} and used to assess TrueWalks~\cite{sousa2023biomedical}, constructed from the STRING database~\cite{STRING2021}. This dataset comprises 440 proteins, 1 024 interacting protein pairs (corresponding to valid examples), and another 1 024 non-interacting protein pairs (corresponding to invalid examples).

\section{Results and Discussion}

From subsections~\ref{sec:linkprediction} to \ref{sec:clustering}, we first present and analyze the effectiveness of the proposed approach when $cl\_phase$ is set to 350. In terms of embedding dimensions, baseline models generate 100-dimensional embeddings, whereas our approach simultaneously trains positive and negative models to create embeddings of 50 dimensions each.
To thoroughly evaluate our approach, we first report performance results for the two tasks (link prediction on Wikidata and triple classification on the GO KG). Next, we investigate the quality of the embeddings on both KGs, Wikidata and GO, by applying clustering metrics to determine how well KG entities of the same semantic type are grouped in the embedding space. 
Finally, subsection~\ref{sec:ablation} presents ablation studies to analyze the impact of varying two hyperparameters ($cl\_phase$ and embedding size).

\begin{table}[!t]
\scriptsize
\caption{Rank-based metrics (MRR, Hits@10, and Hits@1) on the Wikidata KG. The reported metrics are averaged over head and tail predictions. Higher values reflect better results. Baselines are shaded in grey, while bold text highlights the best approach for each metric within each KGE model.}
\label{tab:LP-results}
\begin{tabular}{llrrr}
\toprule
\textbf{KGE Model} &  & \textbf{MRR} & \textbf{Hits@10} & \textbf{Hits@1} \\ \midrule

\multirow{4}{*}{TransE} 
& \cellcolor{gray!10}pos & \cellcolor{gray!10}10.06\%           & \cellcolor{gray!10}18.20\%           & \cellcolor{gray!10}\textbf{5.52\%}  \\
& \cellcolor{gray!10}neg & \cellcolor{gray!10}9.34\%	& \cellcolor{gray!10}17.28\%	& \cellcolor{gray!10}5.08\% \\
& ours (pos)  & 10.15\%           & 18.56\%           & \textbf{5.52\%}  \\
& ours (pos-concat-neg)  & \textbf{10.40\%}  & \textbf{19.32\%}  & 5.36\%           \\ \midrule

\multirow{4}{*}{DistMult} 
& \cellcolor{gray!10}pos & \cellcolor{gray!10}6.86\%            & \cellcolor{gray!10}12.64\%           & \cellcolor{gray!10}3.76\%           \\
& \cellcolor{gray!10}neg & \cellcolor{gray!10}4.052\% & \cellcolor{gray!10}7.480\% & \cellcolor{gray!10}2.240\% \\
& ours (pos)  & 7.90\%            & 17.52\%           & 3.40\%           \\
& ours (pos-concat-neg)  & \textbf{9.63\%}   & \textbf{20.56\%}  & \textbf{4.64\%}  \\ \midrule

\multirow{4}{*}{ComplEx} 
& \cellcolor{gray!10}pos & \cellcolor{gray!10}4.66\%           & \cellcolor{gray!10}10.320\%          & \cellcolor{gray!10}1.840\%          \\
& \cellcolor{gray!10}neg & \cellcolor{gray!10}4.77\%  &  \cellcolor{gray!10}8.60\% & \cellcolor{gray!10}2.96\% \\
& ours (pos) & 8.86\%           & 19.12\%          & 4.16\%          \\
& ours (pos-concat-neg) & \textbf{10.07\%} & \textbf{20.80\%} & \textbf{4.84\%} \\ 

\bottomrule
\end{tabular}
\vspace{-0.3cm}
\end{table}

\subsection{Link Prediction on Wikidata} \label{sec:linkprediction}

Link prediction involves inferring a missing entity in a triple, such as predicting the head entity $h$ given $(r, t)$, or the tail entity $t$ given $(h, r)$. For each test triple $(h, r, t)$, we generate corrupted triples by replacing either the head or the tail entity with every entity in the KG. A scoring function $f(h, r, t)$ is then applied to both the original and corrupted triples. To ensure fair evaluation, we follow previous works and filter out any corrupted triples that appear in the training set, as their inclusion could underestimate the model's performance. Finally, the remaining triples are ranked by their scores to evaluate the model's ability to distinguish valid from invalid triples.

Table~\ref{tab:LP-results} reports the mean reciprocal rank (MRR) and the proportion of valid triples ranked in the top 10 (Hits@10) and top 1 (Hits@1) for both head and tail predictions on Wikidata KG, using three KGE models. 
For each KGE, we compare the baselines trained separately on the positive and negative statements with two variations of our approach: one that uses only the embeddings learned from the positive KG, and another that represents entities by concatenating their embeddings from both the positive and negative KGs.
Across all models, the proposed contrastive learning approach consistently outperforms the baselines for MRR and Hits@10, with the best performance consistently achieved when combining positive and negative embeddings. The improvements are particularly pronounced for DistMult and ComplEx, with MRR values increasing from 6.86\% to 9.63\% or from 4.66\% to 10.07\%. 
The proposed approach narrows the performance gap between simpler models, such as TransE, and more complex ones, like ComplEx and DistMult, so that they converge to similar MRR, Hits@10, and Hits@1 values despite starting with considerable baseline differences.

Interestingly, and somewhat counterintuitively, training a baseline model on a KG with negative statements and then using it to score positive statements yields competitive results, especially for ComplEx, where training on the negative KG outperforms training on the positive KG alone. 
This highlights the value of meaningful negative statements and suggests that, because negative statements are still plausible, training on them helps the model to better distinguish between plausible and implausible triples.

In addition to rank-based metrics, semantic awareness is assessed in Table~\ref{tab:sem-results} using the Sem@K metric~\cite{hubert2023sem}, which measures the proportion of triples that are semantically valid in the first $K$ top-scored triples. Formally, Sem@K is defined as:

\begin{equation}
\text{Sem@}K = \frac{1}{|B|} \sum_{q \in B} \frac{1}{K} \sum_{q' \in S^K_q} compatibility(q, q')
\end{equation}

where for each ground-truth triple $q = (h, r, t)$, the set $S^K_q$ contains the top $K$ candidate triples scored by the KGE model, either by predicting the tail given $(h, r)$ or the head given $(r, t)$. The compatibility function, $compatibility(q, q')$, evaluates whether a candidate triple $q'$ is semantically compatible with its ground-truth $q$. In this work, semantic compatibility means that the predicted entity (head or tail) belongs to the same type as the corresponding entity in the ground-truth triple. For example, given the ground-truth triple $(Mozart, occupation, Composer)$, the prediction $(Mozart, occupation, Teacher)$ is semantically valid since both tails are of type profession, whereas $(Mozart, occupation, Germany)$ is invalid because its tail belongs to the type country.
Table~\ref{tab:sem-results} shows that the proposed dual model approach with contrastive learning yields more semantically valid predictions. The best results for DistMult and ComplEx are obtained when embeddings from both KGs are combined.
These results indicate that the proposed approach not only improves ranking metrics but also leads to more semantically plausible predictions.

\begin{table}[t!]
\scriptsize
\caption{Sem@1 on Wikidata KG, reported separately for predicting the head entity, the tail entity, and their average. Baselines are shaded in grey, while bold text highlights the best approach for each metric within each KGE model.}
\label{tab:sem-results}
\begin{tabular}{llrrr}
\toprule
\textbf{KGE Model} &  & \textbf{Head} & \textbf{Tail} & \textbf{Average} \\ \midrule

\multirow{4}{*}{TransE} 
& \cellcolor{gray!10}pos & \cellcolor{gray!10}53.12\%          & \cellcolor{gray!10}\textbf{62.48\%} & \cellcolor{gray!10}57.80\%\\
& \cellcolor{gray!10}neg & \cellcolor{gray!10}20.96\%          & \cellcolor{gray!10}58.72\%          & \cellcolor{gray!10}39.84\%\\
& ours (pos)  & \textbf{63.36\%} & 57.84\%          & \textbf{60.60\%} \\
& ours (pos-concat-neg)  & 53.04\%          & 62.40\%          & 57.72\% \\ \midrule

\multirow{4}{*}{DistMult} 
& \cellcolor{gray!10}pos &  \cellcolor{gray!10}62.16\%          & \cellcolor{gray!10}40.00\%          & \cellcolor{gray!10}51.08\% \\
& \cellcolor{gray!10}neg &  \cellcolor{gray!10}25.28\%          & \cellcolor{gray!10}46.64\%          & \cellcolor{gray!10}35.96\%  \\
& ours (pos)  &  71.84\%          & 46.48\%          & 59.16\%\\
& ours (pos-concat-neg)  & \textbf{74.40\%} & \textbf{58.96\%} & \textbf{66.68\%}\\ \midrule

\multirow{4}{*}{ComplEx} 
& \cellcolor{gray!10}pos &  \cellcolor{gray!10}76.72\%          & \cellcolor{gray!10}34.08\%          & \cellcolor{gray!10}55.40\%   \\
& \cellcolor{gray!10}neg &  \cellcolor{gray!10}37.44\%          & \cellcolor{gray!10}\textbf{63.20\%} & \cellcolor{gray!10}50.32\% \\
& ours (pos) &  \textbf{80.24\%} & 39.20\%          & 59.72\% \\
& ours (pos-concat-neg) &  74.88\%          & 57.60\%          & \textbf{66.24\%}\\ 

\bottomrule
\end{tabular}
\vspace{-0.3cm}
\end{table}

\subsection{Triple Classification on Gene Ontology} \label{sec:classification}

The triple classification task involves determining whether a given triple $(h, r, t)$ holds or not and can be framed as a binary classification problem. In the context of the GO KG, this task corresponds to predicting whether one protein interacts with another, i.e., whether $(p_1,interacts, p_2)$ is valid. Following the approach in~\cite{sousa2023biomedical}, to predict the relation between a pair of proteins $p_1$ and $p_2$, we first obtain their vector embeddings $v_{p_1}$ and $v_{p_2}$. These embeddings are then combined using the Hadamard product~\cite{horn1990hadamard} to form the pair representation: $ r(p_1, p_2) = v_{p_1} \times v_{p_2}$. The resulting pair representations $r(p_1, p_2)$ serve as input to train a Random Forest classifier~\cite{breiman2001random}, which has been shown to perform very well in PPI prediction~\cite{sousa2023biomedical}.
For the Random Forest classifier, we employ a 5-fold cross-validation strategy. In each fold, the classifier is trained to output predicted labels (0 or 1) for each protein pair. These predicted labels are then used to compute the predictive performance metrics, including precision (Pr), recall (Re), weighted average F-measure (F1), and area under the receiver operating characteristic curve (AUC).

\begin{table}[t!]
\scriptsize
\caption{Median of performance metrics (Pr, Re, F1, and AUC) for PPI prediction using the GO KG, computed over 5-fold cross-validation. Bold text highlights the best performing approach for each metric within each KGE model.}
\label{tab:ppi-results}
\begin{tabular}{llrrrr}
\toprule
\textbf{KGE Model} & & \textbf{Pr} & \textbf{Re} & \textbf{F1} & \textbf{AUC} \\ \midrule

\multirow{4}{*}{TransE} 
& \cellcolor{gray!10}pos  &  \cellcolor{gray!10}58.85\%          & \cellcolor{gray!10}58.68\%          & \cellcolor{gray!10}58.51\%          & \cellcolor{gray!10}62.80\% \\
& \cellcolor{gray!10}neg &\cellcolor{gray!10} 64.07\%          & \cellcolor{gray!10}64.06\%          & \cellcolor{gray!10}64.05\%          & \cellcolor{gray!10}68.49\%  \\
& ours (pos)  & 60.98\%          & 60.98\%          & 60.97\%          & 64.74\% \\
& ours (pos-concat-neg)  &  \textbf{67.34\%} & \textbf{67.32\%} & \textbf{67.31\%} & \textbf{74.13\%}\\ \midrule

\multirow{4}{*}{DistMult} 
& \cellcolor{gray!10}pos & \cellcolor{gray!10}81.67\%          & \cellcolor{gray!10}81.66\%          & \cellcolor{gray!10}81.66\%          & \cellcolor{gray!10}88.91\%  \\
& \cellcolor{gray!10}neg & \cellcolor{gray!10}\textbf{83.70\%} & \cellcolor{gray!10}\textbf{82.89\%} & \cellcolor{gray!10}\textbf{82.78\%} & \cellcolor{gray!10}90.58\%  \\
& ours (pos)  & 77.27\%          & 77.07\%          & 77.03\%          & 86.03\% \\
& ours (pos-concat-neg)  &  83.03\%          & 82.68\%          & 82.64\%          & \textbf{90.93\%}\\ \midrule

\multirow{4}{*}{ComplEx} 
& \cellcolor{gray!10}pos & \cellcolor{gray!10}79.34\%          & \cellcolor{gray!10}79.02\%          & \cellcolor{gray!10}78.94\%          & \cellcolor{gray!10}86.70\%\\
& \cellcolor{gray!10}neg & \cellcolor{gray!10}81.45\%          & \cellcolor{gray!10}80.20\%          & \cellcolor{gray!10}79.99\%          & \cellcolor{gray!10}87.93\%  \\
& ours (pos) &  78.97\%          & 78.97\%          & 78.97\%          & 87.54\% \\
& ours (pos-concat-neg)   &  \textbf{81.91\%} & \textbf{81.91\%} & \textbf{81.91\%} & \textbf{89.55\%}\\
\bottomrule
\end{tabular}
\vspace{-0.3cm}
\end{table}

Table~\ref{tab:ppi-results} reports the median performance metrics for the baselines trained separately on the positive and negative statements, as well as two variations of the proposed approach (one uses only the positive embeddings, and the other concatenates both).
For TransE and ComplEx, the proposed approach leads to improvements across all performance metrics compared to training KGE models solely on positive and negative KGs. For example, TransE shows an improvement in F1 from 58.5\% to 67.3\%. 
Importantly, the differences between the proposed approach and the baselines for both TransE and ComplEx are statistically significant according to the Kruskal–Wallis test~\cite{mckight2010kruskal} with a significance level of 0.05.
The only exception is DistMult, where training solely on the negative KG yields the best performance across most metrics, even outperforming the proposed contrastive learning approach. 
This behavior mirrors the link prediction results, likely because the model learns from plausible negative statements.

Overall, it appears that weaker KGE models, such as TransE, benefit the most from the proposed contrastive learning approach. For instance, TransE achieves the lowest performance scores but exhibits the largest gains when combined with the proposed contrastive learning approach. In contrast, DistMult is the best performing method, and likely due to its already high baseline performance, the proposed approach offers only small improvements.

\subsection{Embeddings Evaluation} \label{sec:clustering}

To gain deeper insights into the distribution of different embeddings, we investigate how entity types are represented using clustering metrics. Rather than performing actual clustering on the data, we use these metrics to assess the degree of natural separation between entities of different types in the embedding space. This provides a proxy for how well the representations obtained with different approaches reflect underlying semantic distinctions.
For Wikidata KG, we consider 21 frequent types, whereas for the GO KG only two classes are used, namely the GO classes and proteins.

Three clustering metrics are employed. The Calinski-Harabasz score~\cite{calinski1974dendrite} measures the ratio of between-cluster dispersion to within-cluster dispersion. Higher values indicate better-defined clusters with greater separation. The Davies-Bouldin score~\cite{davies1979cluster} evaluates the average similarity between each cluster and its most similar peer cluster. Lower values indicate better clustering. The silhouette score~\cite{rousseeuw1987silhouettes} assesses how similar an entity is to its own cluster compared to other clusters. Higher values indicate points well-matched to their own cluster and poorly matched to neighboring clusters.

\begin{figure}[t!]
    \centering
    \begin{subfigure}[b]{0.49\linewidth}
        \centering
        \includegraphics[width=0.3\linewidth]{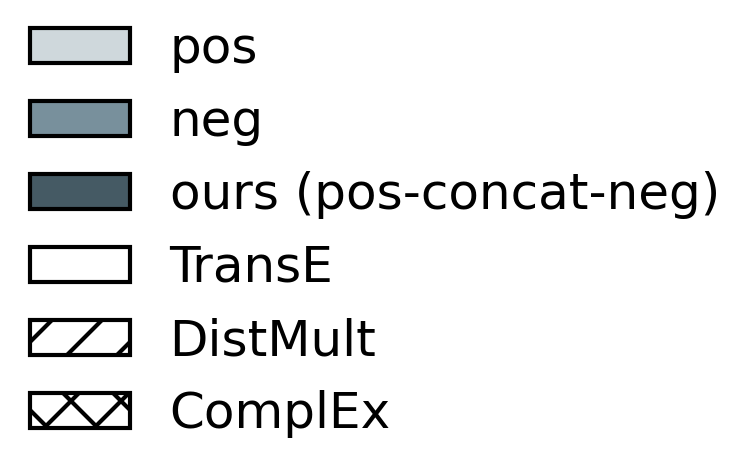}
    \end{subfigure}
    \hfill
    \begin{subfigure}[b]{0.49\linewidth}
        \centering
        \includegraphics[width=0.3\linewidth]{Figures/ClusteringMetrics_label-cr.png}
    \end{subfigure}
    
    \centering
    \begin{subfigure}[b]{0.49\linewidth}
        \centering
        \includegraphics[width=\linewidth]{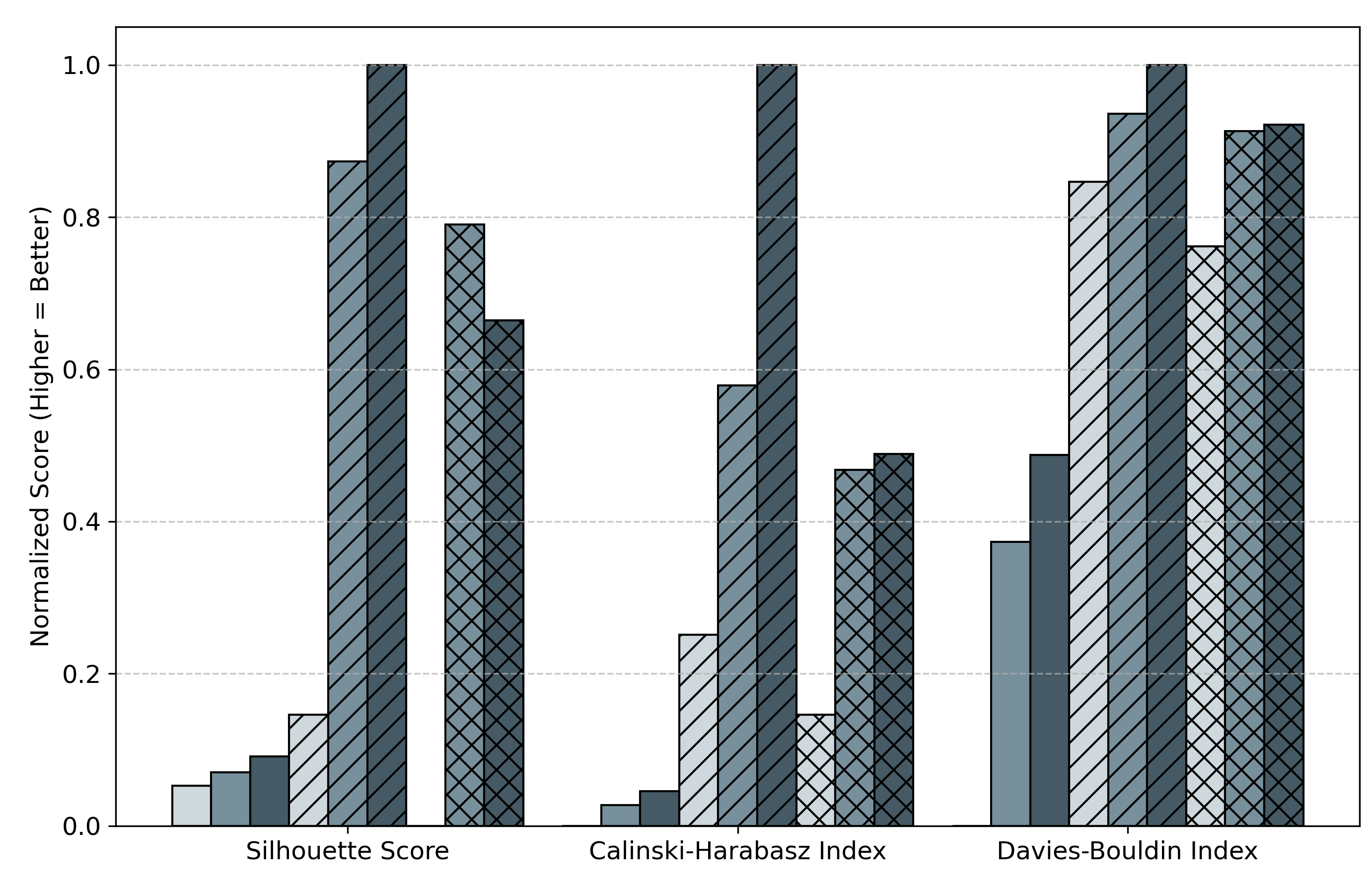}
        \caption{Wikidata KG}
        \label{fig:clusters-wikinegata}
    \end{subfigure}
    \hfill
    \begin{subfigure}[b]{0.49\linewidth}
        \centering
        \includegraphics[width=\linewidth]{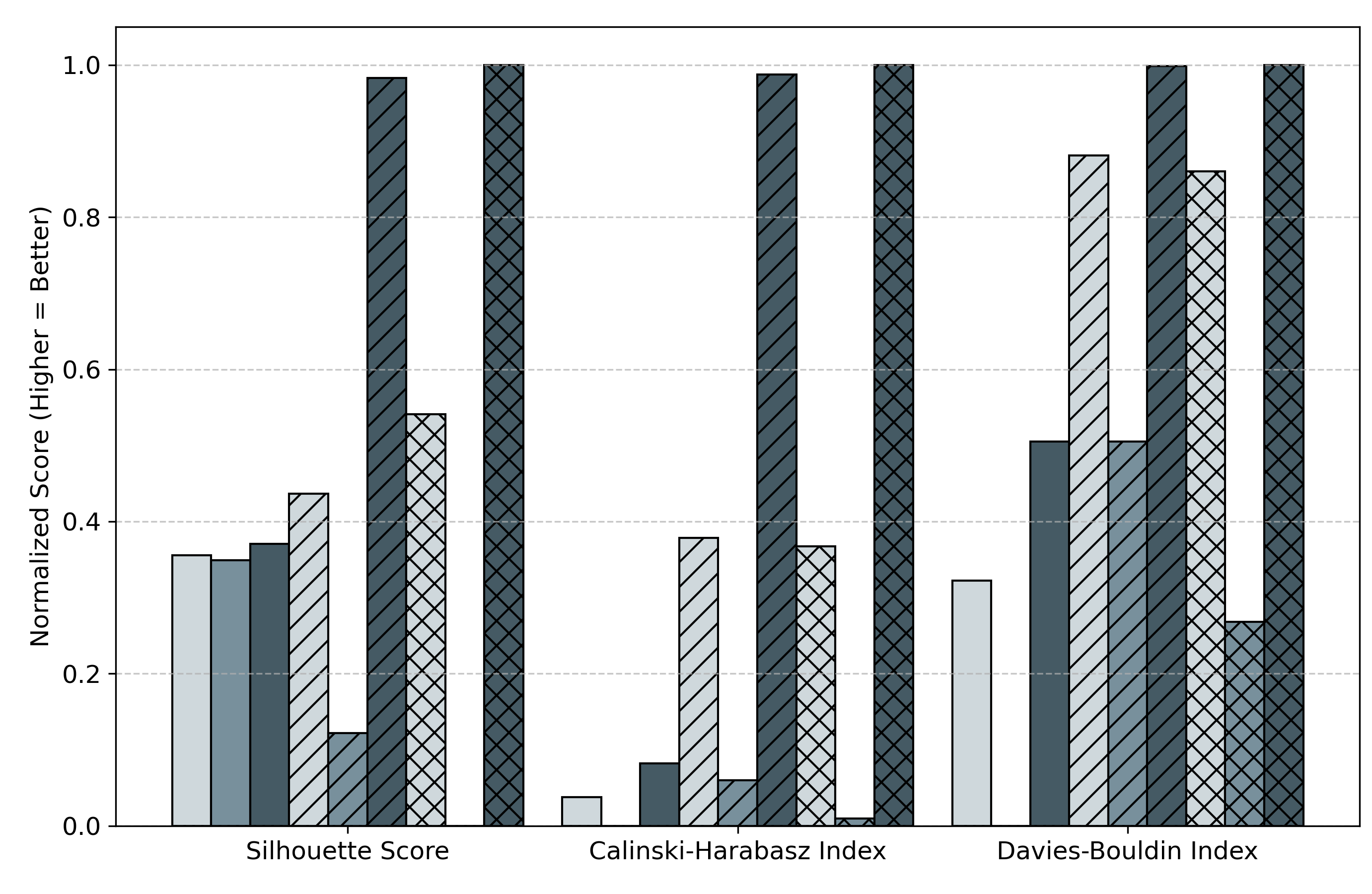}
        \caption{GO KG}
        \label{fig:clusters-ppi}
    \end{subfigure}

    \caption{Bar plots display normalized clustering metric values (Calinski-Harabasz, inverted Davies-Bouldin, and Silhouette) comparing the baselines and the proposed approach (distinguished by colors) across three KGE models (distinguished by bar hatchings). Metric values are normalized to [0, 1] within each metric, with higher values consistently indicating better cluster separation. After normalization, the lowest values for each metric are displayed as 0 in the plots.}
    \label{fig:clusters}
    \vspace{-0.3cm}
\end{figure}

Figure~\ref{fig:clusters} presents bar plots displaying the values of three clustering metrics for baselines trained separately on the positive and negative statements, as well as the proposed approach across three KGE models.
Within each metric, all values are normalized to the [0, 1] range, with the Davies-Bouldin scores inverted to maintain a consistent higher-is-better interpretation. Furthermore, this normalization enables direct visual comparison of the relative effectiveness of each approach. Across all metrics, our approach demonstrates superior clustering performance, highlighting the effectiveness of the contrastive learning strategy in increasing inter-type separability. 

Interestingly, for Wikidata KG, the baseline trained on the negative KG outperforms the baseline trained on the positive KG in terms of cluster separability. 
One possible explanation is that the distribution of relations associated with each entity type in Wikidata differs in the negative KG, which may make the clustering task easier. This distributional bias is not observed in the GO KG, where entities typically participate in the same relations (e.g.,$hasFunction$ $isA$) across both KGs. 
An analysis of the two most frequent entity types ($Person$ and $Business$) in Wikidata revealed that this bias affects business entities, with negative statements centered around a single relation ($structuredAs$), unlike in positive statements.

\subsection{Ablation Studies} \label{sec:ablation}

\begin{figure}[!t]
    \centering
    \begin{subfigure}[b]{0.49\columnwidth}
        \centering
        \includegraphics[width=0.7\linewidth]{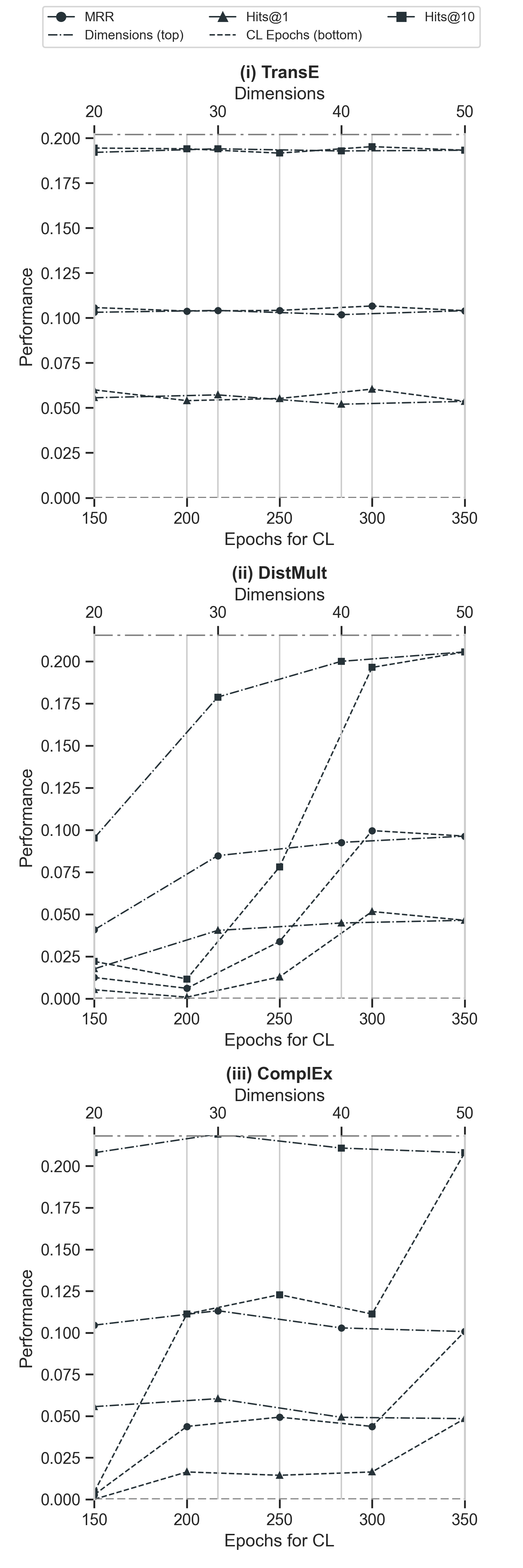}
        \caption{Wikidata KG}
    \end{subfigure}
    \hfill
    \begin{subfigure}[b]{0.49\columnwidth}
        \centering
        \includegraphics[width=0.7\linewidth]{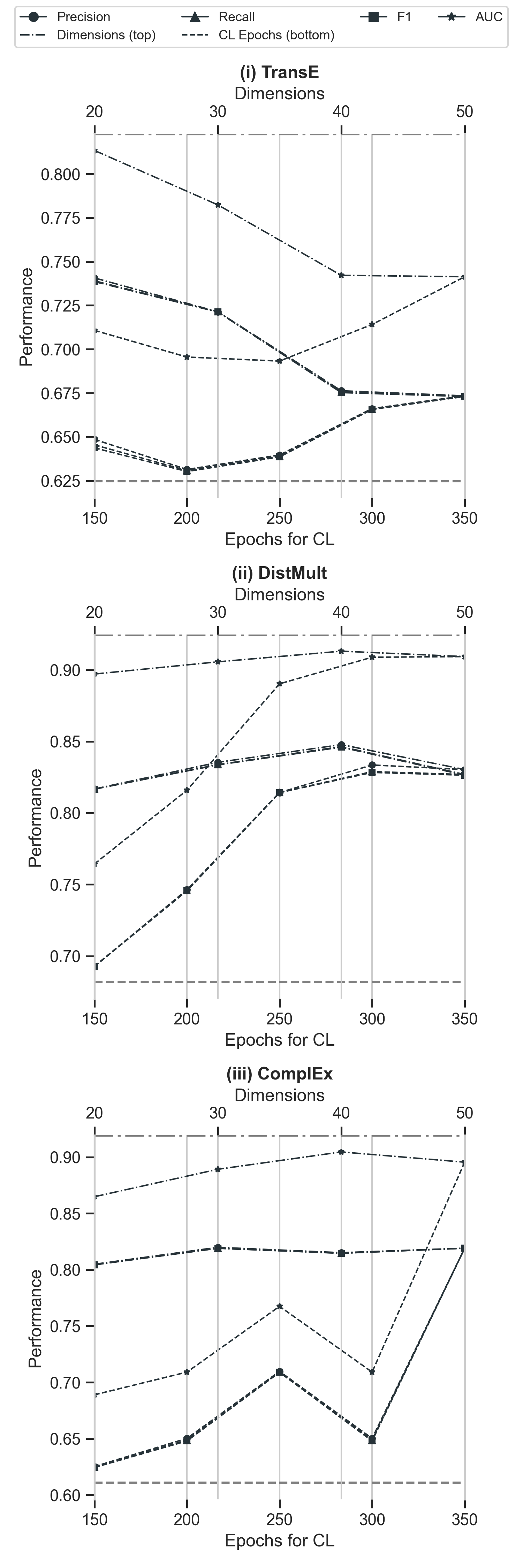}
        \caption{GO KG}
    \end{subfigure}
    \caption{Line plots illustrating the effect of $cl\_phase$ (bottom x-axis) and the embeddings dimensionality (top x-axis) on performance (y-axis) for link prediction on Wikidata KG and triple classification on GO KG.}
    \label{fig:ablation}
    \vspace{-0.3cm}
\end{figure}

We conduct ablation studies to investigate the impact of two hyperparameters of the proposed approach: the dimensionality of the embeddings, and the epoch at which the negative sample generation transitions from negative sampling to a contrastive learning-based strategy ($cl\_phase$). Figure~\ref{fig:ablation} shows performance on both the Wikidata and GO KGs for multiple values of $cl\_phase$ (100, 150, 200, 250, 300 and 350) and embedding dimensions (20, 30, 40, and 50).
The results indicate that, particularly for larger KGs like Wikidata, earlier transitions to contrastive learning (lower $cl\_phase$ values) tend to limit the model's performance, with performance generally increasing as $cl\_phase$ approaches 300. Delaying the transition appears to allow embeddings to stabilize before introducing more challenging contrastive samples.
It is essential to note that during contrastive learning, for each triple in a KG, either the head or the tail entity is replaced with all possible entities from the KG. This incurs  additional computational time per epoch, as the negative samples are dynamically updated rather than being fixed.
Therefore, selecting the appropriate $cl\_phase$ should balance between training time and performance gains.
In terms of embedding dimensionality, the proposed approach demonstrates robustness across different embedding sizes, with relatively stable performance. However, an exception is observed with DistMult on Wikidata KG, where larger dimensions lead to substantial performance gain. 

\section{Conclusion}

KGE methods are increasingly applied in a wide array of real-world applications across multiple domains. However, existing KGE models are not designed to handle meaningful negative statements that are explicitly defined, despite their recognized importance in knowledge representation. 
To overcome this limitation, we propose a novel approach that combines dual-model training with an adapted negative sampling mechanism grounded in contrastive learning.
Our empirical evaluation demonstrates that the proposed approach outperforms state-of-the-art KGE models on both general-purpose (Wikidata) and domain-specific (GO) KGs, yielding substantial gains in link prediction and triple classification tasks.
Given its flexible design, the proposed approach can be easily incorporated into any KGE model that employs a scoring function. Moreover, it offers generalizability across different KGs and predictive tasks.   

For future work, several directions can be explored to improve the proposed approach.
First, in contrastive learning, for each triple in the KG, either the head or tail is paired with all KG entities. While effective, this approach is not scalable for very large KGs. Future work could investigate more efficient strategies, such as random sampling from the set of entities or selecting entities based on their embedding proximity, to substantially reduce computational cost. Second, further improvements could be achieved through systematic hyperparameter optimization, which may lead to additional performance gains. Finally, the current implementation trains the two models synchronously, applying the same number of initial training epochs to both the positive and negative KGE models. However, an asynchronous training strategy, allowing different numbers of epochs for each model, could be explored in the future.

\vspace{-0.3cm}
\bibliographystyle{ACM-Reference-Format}
\bibliography{mybibliography}


\begin{thebibliography}{40}


\ifx \showCODEN    \undefined \def \showCODEN     #1{\unskip}     \fi
\ifx \showISBNx    \undefined \def \showISBNx     #1{\unskip}     \fi
\ifx \showISBNxiii \undefined \def \showISBNxiii  #1{\unskip}     \fi
\ifx \showISSN     \undefined \def \showISSN      #1{\unskip}     \fi
\ifx \showLCCN     \undefined \def \showLCCN      #1{\unskip}     \fi
\ifx \shownote     \undefined \def \shownote      #1{#1}          \fi
\ifx \showarticletitle \undefined \def \showarticletitle #1{#1}   \fi
\ifx \showURL      \undefined \def \showURL       {\relax}        \fi
\providecommand\bibfield[2]{#2}
\providecommand\bibinfo[2]{#2}
\providecommand\natexlab[1]{#1}
\providecommand\showeprint[2][]{arXiv:#2}

\bibitem[Arnaout et~al\mbox{.}(2023)]%
        {arnaout2023uncommonsense}
\bibfield{author}{\bibinfo{person}{Hiba Arnaout}, \bibinfo{person}{Tuan-Phong Nguyen}, \bibinfo{person}{Simon Razniewski}, {and} \bibinfo{person}{Gerhard Weikum}.} \bibinfo{year}{2023}\natexlab{}.
\newblock \showarticletitle{UnCommonSense in Action! Informative Negations for Commonsense Knowledge Bases}. In \bibinfo{booktitle}{\emph{ACM International Conference on Web Search and Data Mining}}. \bibinfo{pages}{1120--1123}.
\newblock


\bibitem[Arnaout et~al\mbox{.}(2021a)]%
        {arnaout_negative_2021}
\bibfield{author}{\bibinfo{person}{Hiba Arnaout}, \bibinfo{person}{Simon Razniewski}, \bibinfo{person}{Gerhard Weikum}, {and} \bibinfo{person}{Jeff~Z. Pan}.} \bibinfo{year}{2021}\natexlab{a}.
\newblock \showarticletitle{{Negative knowledge for open-world wikidata}}. In \bibinfo{booktitle}{\emph{{The Web Conference}}}. \bibinfo{pages}{544--551}.
\newblock


\bibitem[Arnaout et~al\mbox{.}(2021b)]%
        {arnaout2021wikinegata}
\bibfield{author}{\bibinfo{person}{Hiba Arnaout}, \bibinfo{person}{Simon Razniewski}, \bibinfo{person}{Gerhard Weikum}, {and} \bibinfo{person}{Jeff~Z Pan}.} \bibinfo{year}{2021}\natexlab{b}.
\newblock \showarticletitle{Wikinegata: a knowledge base with interesting negative statements}.
\newblock \bibinfo{journal}{\emph{Proceedings of the VLDB Endowment}} \bibinfo{volume}{14}, \bibinfo{number}{12} (\bibinfo{year}{2021}), \bibinfo{pages}{2807--2810}.
\newblock


\bibitem[Bordes et~al\mbox{.}(2013)]%
        {bordes2013translating}
\bibfield{author}{\bibinfo{person}{Antoine Bordes}, \bibinfo{person}{Nicolas Usunier}, \bibinfo{person}{Alberto Garcia-Duran}, \bibinfo{person}{Jason Weston}, {and} \bibinfo{person}{Oksana Yakhnenko}.} \bibinfo{year}{2013}\natexlab{}.
\newblock \showarticletitle{Translating embeddings for modeling multi-relational data}.
\newblock \bibinfo{journal}{\emph{Advances in Neural Information Processing Systems}}  \bibinfo{volume}{26} (\bibinfo{year}{2013}).
\newblock


\bibitem[Breiman(2001)]%
        {breiman2001random}
\bibfield{author}{\bibinfo{person}{Leo Breiman}.} \bibinfo{year}{2001}\natexlab{}.
\newblock \showarticletitle{Random forests}.
\newblock \bibinfo{journal}{\emph{Machine learning}}  \bibinfo{volume}{45} (\bibinfo{year}{2001}), \bibinfo{pages}{5--32}.
\newblock


\bibitem[Cali{\'n}ski and Harabasz(1974)]%
        {calinski1974dendrite}
\bibfield{author}{\bibinfo{person}{Tadeusz Cali{\'n}ski} {and} \bibinfo{person}{Jerzy Harabasz}.} \bibinfo{year}{1974}\natexlab{}.
\newblock \showarticletitle{A dendrite method for cluster analysis}.
\newblock \bibinfo{journal}{\emph{Communications in Statistics-theory and Methods}} \bibinfo{volume}{3}, \bibinfo{number}{1} (\bibinfo{year}{1974}), \bibinfo{pages}{1--27}.
\newblock


\bibitem[Davies and Bouldin(1979)]%
        {davies1979cluster}
\bibfield{author}{\bibinfo{person}{David~L Davies} {and} \bibinfo{person}{Donald~W Bouldin}.} \bibinfo{year}{1979}\natexlab{}.
\newblock \showarticletitle{A cluster separation measure}.
\newblock \bibinfo{journal}{\emph{IEEE Transactions on Pattern Analysis and Machine Intelligence}} \bibinfo{number}{2} (\bibinfo{year}{1979}), \bibinfo{pages}{224--227}.
\newblock


\bibitem[Doherty et~al\mbox{.}(2000)]%
        {doherty2000efficient}
\bibfield{author}{\bibinfo{person}{Patrick Doherty}, \bibinfo{person}{Witold Lukaszewicz}, {and} \bibinfo{person}{Andrzej Szalas}.} \bibinfo{year}{2000}\natexlab{}.
\newblock \showarticletitle{{Efficient reasoning using the local closed-world assumption}}. In \bibinfo{booktitle}{\emph{International Conference on Artificial Intelligence: Methodology, Systems, and Applications}}. \bibinfo{pages}{49--58}.
\newblock


\bibitem[Gal{\'a}rraga et~al\mbox{.}(2013)]%
        {galarraga2013amie}
\bibfield{author}{\bibinfo{person}{Luis~Antonio Gal{\'a}rraga}, \bibinfo{person}{Christina Teflioudi}, \bibinfo{person}{Katja Hose}, {and} \bibinfo{person}{Fabian Suchanek}.} \bibinfo{year}{2013}\natexlab{}.
\newblock \showarticletitle{{AMIE: association rule mining under incomplete evidence in ontological knowledge bases}}. In \bibinfo{booktitle}{\emph{International Conference on World Wide Web}}. \bibinfo{pages}{413--422}.
\newblock


\bibitem[{GO Consortium}(2021)]%
        {gene2021gene}
\bibfield{author}{\bibinfo{person}{{GO Consortium}}.} \bibinfo{year}{2021}\natexlab{}.
\newblock \showarticletitle{{The Gene Ontology resource: enriching a GOld mine}}.
\newblock \bibinfo{journal}{\emph{Nucleic Acids Research}} \bibinfo{volume}{49}, \bibinfo{number}{D1} (\bibinfo{year}{2021}), \bibinfo{pages}{D325--D334}.
\newblock


\bibitem[Guo et~al\mbox{.}(2020)]%
        {guo2020survey}
\bibfield{author}{\bibinfo{person}{Qingyu Guo}, \bibinfo{person}{Fuzhen Zhuang}, \bibinfo{person}{Chuan Qin}, \bibinfo{person}{Hengshu Zhu}, \bibinfo{person}{Xing Xie}, \bibinfo{person}{Hui Xiong}, {and} \bibinfo{person}{Qing He}.} \bibinfo{year}{2020}\natexlab{}.
\newblock \showarticletitle{A survey on knowledge graph-based recommender systems}.
\newblock \bibinfo{journal}{\emph{IEEE Transactions on Knowledge and Data Engineering}} \bibinfo{volume}{34}, \bibinfo{number}{8} (\bibinfo{year}{2020}), \bibinfo{pages}{3549--3568}.
\newblock


\bibitem[Hogan et~al\mbox{.}(2021)]%
        {hogan2021knowledge}
\bibfield{author}{\bibinfo{person}{Aidan Hogan}, \bibinfo{person}{Eva Blomqvist}, \bibinfo{person}{Michael Cochez}, \bibinfo{person}{Claudia d’Amato}, \bibinfo{person}{Gerard~de Melo}, \bibinfo{person}{Claudio Gutierrez}, \bibinfo{person}{Sabrina Kirrane}, {et~al\mbox{.}}} \bibinfo{year}{2021}\natexlab{}.
\newblock \showarticletitle{{Knowledge Graphs}}.
\newblock \bibinfo{journal}{\emph{{ACM Computing Surveys}}} \bibinfo{volume}{54}, \bibinfo{number}{4} (\bibinfo{year}{2021}), \bibinfo{pages}{1--37}.
\newblock


\bibitem[Horn(1990)]%
        {horn1990hadamard}
\bibfield{author}{\bibinfo{person}{Roger~A Horn}.} \bibinfo{year}{1990}\natexlab{}.
\newblock \showarticletitle{The hadamard product}. In \bibinfo{booktitle}{\emph{Proceedings of Symposia in Applied Mathematics}}, Vol.~\bibinfo{volume}{40}. \bibinfo{pages}{87--169}.
\newblock


\bibitem[Huang et~al\mbox{.}(2019a)]%
        {huang2019signed}
\bibfield{author}{\bibinfo{person}{Junjie Huang}, \bibinfo{person}{Huawei Shen}, \bibinfo{person}{Liang Hou}, {and} \bibinfo{person}{Xueqi Cheng}.} \bibinfo{year}{2019}\natexlab{a}.
\newblock \showarticletitle{Signed graph attention networks}. In \bibinfo{booktitle}{\emph{International Conference on Artificial Neural Networks}}. Springer, \bibinfo{pages}{566--577}.
\newblock


\bibitem[Huang et~al\mbox{.}(2023)]%
        {huang2023negative}
\bibfield{author}{\bibinfo{person}{Junjie Huang}, \bibinfo{person}{Ruobing Xie}, \bibinfo{person}{Qi Cao}, \bibinfo{person}{Huawei Shen}, \bibinfo{person}{Shaoliang Zhang}, \bibinfo{person}{Feng Xia}, {and} \bibinfo{person}{Xueqi Cheng}.} \bibinfo{year}{2023}\natexlab{}.
\newblock \showarticletitle{Negative can be positive: Signed graph neural networks for recommendation}.
\newblock \bibinfo{journal}{\emph{Information processing and management}} \bibinfo{volume}{60}, \bibinfo{number}{4} (\bibinfo{year}{2023}), \bibinfo{pages}{103403}.
\newblock


\bibitem[Huang et~al\mbox{.}(2019b)]%
        {huang2019knowledge}
\bibfield{author}{\bibinfo{person}{Xiao Huang}, \bibinfo{person}{Jingyuan Zhang}, \bibinfo{person}{Dingcheng Li}, {and} \bibinfo{person}{Ping Li}.} \bibinfo{year}{2019}\natexlab{b}.
\newblock \showarticletitle{Knowledge graph embedding based question answering}. In \bibinfo{booktitle}{\emph{ACM International Conference on Web Search and Data Mining}}. \bibinfo{pages}{105--113}.
\newblock


\bibitem[Hubert et~al\mbox{.}(2023)]%
        {hubert2023sem}
\bibfield{author}{\bibinfo{person}{Nicolas Hubert}, \bibinfo{person}{Pierre Monnin}, \bibinfo{person}{Armelle Brun}, {and} \bibinfo{person}{Davy Monticolo}.} \bibinfo{year}{2023}\natexlab{}.
\newblock \showarticletitle{Sem@ K: Is my knowledge graph embedding model semantic-aware?}
\newblock \bibinfo{journal}{\emph{Semantic Web}} \bibinfo{volume}{14}, \bibinfo{number}{6} (\bibinfo{year}{2023}), \bibinfo{pages}{1273--1309}.
\newblock


\bibitem[Hubert et~al\mbox{.}(2024)]%
        {hubert2024treat}
\bibfield{author}{\bibinfo{person}{Nicolas Hubert}, \bibinfo{person}{Pierre Monnin}, \bibinfo{person}{Armelle Brun}, {and} \bibinfo{person}{Davy Monticolo}.} \bibinfo{year}{2024}\natexlab{}.
\newblock \showarticletitle{Treat different negatives differently: Enriching loss functions with domain and range constraints for link prediction}. In \bibinfo{booktitle}{\emph{Extended Semantic Web Conference}}. \bibinfo{pages}{22--40}.
\newblock


\bibitem[Kotnis and Nastase(2017)]%
        {kotnis2017analysis}
\bibfield{author}{\bibinfo{person}{Bhushan Kotnis} {and} \bibinfo{person}{Vivi Nastase}.} \bibinfo{year}{2017}\natexlab{}.
\newblock \showarticletitle{Analysis of the impact of negative sampling on link prediction in knowledge graphs}.
\newblock \bibinfo{journal}{\emph{arXiv:1708.06816}} (\bibinfo{year}{2017}).
\newblock


\bibitem[Krompa{\ss} et~al\mbox{.}(2015)]%
        {krompass2015type}
\bibfield{author}{\bibinfo{person}{Denis Krompa{\ss}}, \bibinfo{person}{Stephan Baier}, {and} \bibinfo{person}{Volker Tresp}.} \bibinfo{year}{2015}\natexlab{}.
\newblock \showarticletitle{Type-constrained representation learning in knowledge graphs}. In \bibinfo{booktitle}{\emph{International Semantic Web Conference}}. Springer, \bibinfo{pages}{640--655}.
\newblock


\bibitem[Liu et~al\mbox{.}(2023)]%
        {liu2023pane}
\bibfield{author}{\bibinfo{person}{Ziyang Liu}, \bibinfo{person}{Chaokun Wang}, \bibinfo{person}{Jingcao Xu}, \bibinfo{person}{Cheng Wu}, \bibinfo{person}{Kai Zheng}, \bibinfo{person}{Yang Song}, \bibinfo{person}{Na Mou}, {and} \bibinfo{person}{Kun Gai}.} \bibinfo{year}{2023}\natexlab{}.
\newblock \showarticletitle{{PANE-GNN}: Unifying positive and negative edges in graph neural networks for recommendation}.
\newblock \bibinfo{journal}{\emph{arXiv:2306.04095}} (\bibinfo{year}{2023}).
\newblock


\bibitem[Margatina et~al\mbox{.}(2021)]%
        {margatina2021active}
\bibfield{author}{\bibinfo{person}{Katerina Margatina}, \bibinfo{person}{Giorgos Vernikos}, \bibinfo{person}{Lo{\"\i}c Barrault}, {and} \bibinfo{person}{Nikolaos Aletras}.} \bibinfo{year}{2021}\natexlab{}.
\newblock \showarticletitle{Active learning by acquiring contrastive examples}.
\newblock \bibinfo{journal}{\emph{arXiv:2109.03764}} (\bibinfo{year}{2021}).
\newblock


\bibitem[McKight and Najab(2010)]%
        {mckight2010kruskal}
\bibfield{author}{\bibinfo{person}{Patrick~E McKight} {and} \bibinfo{person}{Julius Najab}.} \bibinfo{year}{2010}\natexlab{}.
\newblock \showarticletitle{Kruskal-wallis test}.
\newblock \bibinfo{journal}{\emph{The corsini encyclopedia of psychology}} (\bibinfo{year}{2010}), \bibinfo{pages}{1--1}.
\newblock


\bibitem[Portisch et~al\mbox{.}(2022)]%
        {portisch2022knowledge}
\bibfield{author}{\bibinfo{person}{Jan Portisch}, \bibinfo{person}{Nicolas Heist}, {and} \bibinfo{person}{Heiko Paulheim}.} \bibinfo{year}{2022}\natexlab{}.
\newblock \showarticletitle{{Knowledge graph embedding for data mining vs. knowledge graph embedding for link prediction--two sides of the same coin?}}
\newblock \bibinfo{journal}{\emph{Semantic Web}} (\bibinfo{year}{2022}), \bibinfo{pages}{1--24}.
\newblock


\bibitem[Ristoski and Paulheim(2016)]%
        {ristoski2016rdf2vec}
\bibfield{author}{\bibinfo{person}{Petar Ristoski} {and} \bibinfo{person}{Heiko Paulheim}.} \bibinfo{year}{2016}\natexlab{}.
\newblock \showarticletitle{{RDF2V}ec: {RDF} graph embeddings for data mining}. In \bibinfo{booktitle}{\emph{International Semantic Web Conference}}. Springer, \bibinfo{pages}{498--514}.
\newblock


\bibitem[Rossi et~al\mbox{.}(2021)]%
        {rossi2021knowledge}
\bibfield{author}{\bibinfo{person}{Andrea Rossi}, \bibinfo{person}{Denilson Barbosa}, \bibinfo{person}{Donatella Firmani}, \bibinfo{person}{Antonio Matinata}, {and} \bibinfo{person}{Paolo Merialdo}.} \bibinfo{year}{2021}\natexlab{}.
\newblock \showarticletitle{Knowledge graph embedding for link prediction: A comparative analysis}.
\newblock \bibinfo{journal}{\emph{ACM Transactions on Knowledge Discovery from Data}} \bibinfo{volume}{15}, \bibinfo{number}{2} (\bibinfo{year}{2021}), \bibinfo{pages}{1--49}.
\newblock


\bibitem[Rousseeuw(1987)]%
        {rousseeuw1987silhouettes}
\bibfield{author}{\bibinfo{person}{Peter~J Rousseeuw}.} \bibinfo{year}{1987}\natexlab{}.
\newblock \showarticletitle{{Silhouettes: a graphical aid to the interpretation and validation of cluster analysis}}.
\newblock \bibinfo{journal}{\emph{{Journal of Computational and Applied Mathematics}}}  \bibinfo{volume}{20} (\bibinfo{year}{1987}), \bibinfo{pages}{53--65}.
\newblock


\bibitem[Shaban and Paulheim(2024)]%
        {shaban2024snape}
\bibfield{author}{\bibinfo{person}{Ali Shaban} {and} \bibinfo{person}{Heiko Paulheim}.} \bibinfo{year}{2024}\natexlab{}.
\newblock \showarticletitle{SnapE--training snapshot ensembles of link prediction models}. In \bibinfo{booktitle}{\emph{International Semantic Web Conference}}. Springer, \bibinfo{pages}{3--22}.
\newblock


\bibitem[Shenoy et~al\mbox{.}(2022)]%
        {shenoy2022study}
\bibfield{author}{\bibinfo{person}{Kartik Shenoy}, \bibinfo{person}{Filip Ilievski}, \bibinfo{person}{Daniel Garijo}, \bibinfo{person}{Daniel Schwabe}, {and} \bibinfo{person}{Pedro Szekely}.} \bibinfo{year}{2022}\natexlab{}.
\newblock \showarticletitle{A study of the quality of Wikidata}.
\newblock \bibinfo{journal}{\emph{Journal of Web Semantics}}  \bibinfo{volume}{72} (\bibinfo{year}{2022}), \bibinfo{pages}{100679}.
\newblock


\bibitem[Socher et~al\mbox{.}(2013)]%
        {socher2013reasoning}
\bibfield{author}{\bibinfo{person}{Richard Socher}, \bibinfo{person}{Danqi Chen}, \bibinfo{person}{Christopher~D Manning}, {and} \bibinfo{person}{Andrew Ng}.} \bibinfo{year}{2013}\natexlab{}.
\newblock \showarticletitle{Reasoning with neural tensor networks for knowledge base completion}.
\newblock \bibinfo{journal}{\emph{Advances in Neural Information Processing Systems}}  \bibinfo{volume}{26} (\bibinfo{year}{2013}).
\newblock


\bibitem[Sousa et~al\mbox{.}(2023b)]%
        {sousa2023biomedical}
\bibfield{author}{\bibinfo{person}{Rita~T Sousa}, \bibinfo{person}{Sara Silva}, \bibinfo{person}{Heiko Paulheim}, {and} \bibinfo{person}{Catia Pesquita}.} \bibinfo{year}{2023}\natexlab{b}.
\newblock \showarticletitle{Biomedical knowledge graph embeddings with negative statements}. In \bibinfo{booktitle}{\emph{International Semantic Web Conference}}. Springer, \bibinfo{pages}{428--446}.
\newblock


\bibitem[Sousa et~al\mbox{.}(2020)]%
        {sousa2020evolving}
\bibfield{author}{\bibinfo{person}{Rita~T Sousa}, \bibinfo{person}{Sara Silva}, {and} \bibinfo{person}{Catia Pesquita}.} \bibinfo{year}{2020}\natexlab{}.
\newblock \showarticletitle{Evolving knowledge graph similarity for supervised learning in complex biomedical domains}.
\newblock \bibinfo{journal}{\emph{BMC Bioinformatics}} \bibinfo{volume}{21}, \bibinfo{number}{1} (\bibinfo{year}{2020}), \bibinfo{pages}{1--19}.
\newblock


\bibitem[Sousa et~al\mbox{.}(2023a)]%
        {sousa2023benchmark}
\bibfield{author}{\bibinfo{person}{Rita~T. Sousa}, \bibinfo{person}{Sara Silva}, {and} \bibinfo{person}{Catia Pesquita}.} \bibinfo{year}{2023}\natexlab{a}.
\newblock \showarticletitle{Benchmark datasets for biomedical knowledge graphs with negative statements}. In \bibinfo{booktitle}{\emph{Workshop on Semantic Web Solutions for Large-scale Biomedical Data Analytics co-located with Extended Semantic Web Conference}}.
\newblock


\bibitem[Szklarczyk(2020)]%
        {STRING2021}
\bibfield{author}{\bibinfo{person}{Damian et~al. Szklarczyk}.} \bibinfo{year}{2020}\natexlab{}.
\newblock \showarticletitle{{The STRING database in 2021: customizable protein–protein networks, and functional characterization of user-uploaded gene/measurement sets}}.
\newblock \bibinfo{journal}{\emph{Nucleic Acids Research}} \bibinfo{volume}{49}, \bibinfo{number}{D1} (\bibinfo{date}{11} \bibinfo{year}{2020}), \bibinfo{pages}{D605--D612}.
\newblock
\showISSN{0305-1048}


\bibitem[Trouillon et~al\mbox{.}(2016)]%
        {trouillon2016complex}
\bibfield{author}{\bibinfo{person}{Th\'{e}o Trouillon}, \bibinfo{person}{Johannes Welbl}, \bibinfo{person}{Sebastian Riedel}, \bibinfo{person}{\'{E}ric Gaussier}, {and} \bibinfo{person}{Guillaume Bouchard}.} \bibinfo{year}{2016}\natexlab{}.
\newblock \showarticletitle{{Complex Embeddings for Simple Link Prediction}}. In \bibinfo{booktitle}{\emph{International Conference on International Conference on Machine Learning}}, Vol.~\bibinfo{volume}{48}. \bibinfo{pages}{2071--2080}.
\newblock


\bibitem[Vrande{\v{c}}i{\'c} and Kr{\"o}tzsch(2014)]%
        {vrandevcic2014wikidata}
\bibfield{author}{\bibinfo{person}{Denny Vrande{\v{c}}i{\'c}} {and} \bibinfo{person}{Markus Kr{\"o}tzsch}.} \bibinfo{year}{2014}\natexlab{}.
\newblock \showarticletitle{{Wikidata: a free collaborative knowledgebase}}.
\newblock \bibinfo{journal}{\emph{{Communications of the ACM}}} \bibinfo{volume}{57}, \bibinfo{number}{10} (\bibinfo{year}{2014}), \bibinfo{pages}{78--85}.
\newblock


\bibitem[Wang et~al\mbox{.}(2017)]%
        {wang2017knowledge}
\bibfield{author}{\bibinfo{person}{Quan Wang}, \bibinfo{person}{Zhendong Mao}, \bibinfo{person}{Bin Wang}, {and} \bibinfo{person}{Li Guo}.} \bibinfo{year}{2017}\natexlab{}.
\newblock \showarticletitle{Knowledge graph embedding: A survey of approaches and applications}.
\newblock \bibinfo{journal}{\emph{IEEE Transactions on Knowledge and Data Engineering}} \bibinfo{volume}{29}, \bibinfo{number}{12} (\bibinfo{year}{2017}), \bibinfo{pages}{2724--2743}.
\newblock


\bibitem[Wang et~al\mbox{.}(2014)]%
        {wang2014knowledge}
\bibfield{author}{\bibinfo{person}{Zhen Wang}, \bibinfo{person}{Jianwen Zhang}, \bibinfo{person}{Jianlin Feng}, {and} \bibinfo{person}{Zheng Chen}.} \bibinfo{year}{2014}\natexlab{}.
\newblock \showarticletitle{{Knowledge Graph Embedding by Translating on Hyperplanes}}. In \bibinfo{booktitle}{\emph{AAAI Conference on Artificial Intelligence}}. \bibinfo{pages}{1112--1119}.
\newblock


\bibitem[Warwick~Vesztrocy and Dessimoz(2020)]%
        {vesztrocy2020benchmarking}
\bibfield{author}{\bibinfo{person}{Alex Warwick~Vesztrocy} {and} \bibinfo{person}{Christophe Dessimoz}.} \bibinfo{year}{2020}\natexlab{}.
\newblock \showarticletitle{{{B}enchmarking {G}ene {O}ntology function predictions using negative annotations}}.
\newblock \bibinfo{journal}{\emph{Bioinformatics}} \bibinfo{volume}{36}, \bibinfo{number}{Supplement\_1} (\bibinfo{date}{07} \bibinfo{year}{2020}), \bibinfo{pages}{i210--i218}.
\newblock
\showISSN{1367-4803}


\bibitem[Yang et~al\mbox{.}(2015)]%
        {yang2014embedding}
\bibfield{author}{\bibinfo{person}{Bishan Yang}, \bibinfo{person}{Scott Wen-tau Yih}, \bibinfo{person}{Xiaodong He}, \bibinfo{person}{Jianfeng Gao}, {and} \bibinfo{person}{Li Deng}.} \bibinfo{year}{2015}\natexlab{}.
\newblock \showarticletitle{Embedding Entities and Relations for Learning and Inference in Knowledge Bases}. In \bibinfo{booktitle}{\emph{International Conference on Learning Representations}}.
\newblock


\end{thebibliography}

\appendix
\section{Hyperparameters for Embedding Models}

For our experiments, we adapt the OpenKE library\footnote{https://github.com/thunlp/OpenKE/tree/OpenKE-Tensorflow1.0}. All models (TransE, DistMult, ComplEx) are trained for 400 epochs with 100 batches, an entity negative sampling rate of 1, and a relation negative rate of 0. TransE uses SGD for optimization, while DistMult and ComplEx use Adagrad.
All remaining hyperparameters were left at their default values provided by the OpenKE library.

\end{document}